\documentclass[runningheads]{llncs}
\usepackage{templates/mainpreambel}

\newcommand{\OR}[1]{\color{red}\textbf{[OR]}\color{black}\unskip}
\newcommand{\optional}[1]{\emph{#1}}

\newcommand{\cprob}[2]{p\left(#1 \mid #2\right)}
\newcommand{\prob}[1]{p\left(#1\right)}

\newcommand{\needscite}[1]{
    \color{purple}
    \ifthenelse{\equal{#1}{}}{\textsuperscript{CITE}}{\textsuperscript{CITE #1}}
    \color{black}\ignorespacesafterend\noindent
}

\newenvironment{optional*}[1][]
{
    \noindent
    \color{blue}
    \textit{OPTIONAL\ifthenelse{\isempty{#1}}{}{#1}:}
}
{
    \color{black}
}

\newcommand{\ModelRNG}{Random-Search Generator }
\newcommand{\ModelCBG}{Casebased-Search Generator }

\newcommand{\NumEvoCombinations}{135 }

\makeglossaries
\DeclareDocumentCommand{\newdualentry}{ O{} O{} m m m m } {
        \newglossaryentry{gls-#3}{
                name={#5},
                text={#5\glsadd{#3}},
                description={#6},
                #1
        }
        \newacronym[see={[see:]{gls-#3}},#2]{#3}{#4}{#5\glsadd{gls-#3}}
}

\newglossaryentry{event}
{
        name=process event,
        description={Also called activities. A discrete step in the process.}
}

\newglossaryentry{instance}
{
        name=process instance,
        description={Also called case. A collection of activities that belong to a common entity that is produced by the process.}
}

\newglossaryentry{log}
{
        name=event log,
        description={A collection of event data, that's produced by the process. They are the main input of every process mining venture.}
}
\newglossaryentry{trueprocess}
{
        name=\emph{true} process,
        description={The true data generating process which remains unknown. The term contrasts a modelled process, which seeks to approximate the \emph{true} process' behaviour.}
}

\newglossaryentry{drift}
{
        name=Concept Drift,
        description={The phenomenon of changing processes over time. The changes may occur gradually or sudden and be irregular or recurrent.}
}
\newglossaryentry{rashomon}
{
        name=Rashommon Effect,
        description={Rashomon is a classic Japanese movie in which multiple witnesses tell a different equally valid story about the murder of a samurai. Although, each story acts as a valid explanation, they contradict each other. The same effect may apply to equally valid counterfactuals.}
}
\newglossaryentry{DL}
{
        name=Deep Learning,
        description={A sub-discipline of machine learning which focuses on neural networks as primary tool. The discipline emphasises the research and development of neural network architectures. Data preprocessing and feature engineering play a secondary role.}
}
\newglossaryentry{causalinference}
{
        name=Causal Inference,
        description={A discipline which seeks to incoporate causal relations to model phenomena in the real world.}
}
\newglossaryentry{minkowski}
{
        name=minkowsky distance,
        description={A family of metrics that are defined by an order $p$. The formula follows $D(X, Y)=\left(\sum_{i=1}^{n}\left|x_{i}-y_{i}\right|^{p}\right)^{\frac{1}{p}}$}
}
\newglossaryentry{damerau_levenshtein}
{
        name=Damerau-Levenshstein distance,
        description={A distance function, which allows to compute the dissimilarity of two discrete sequences based on the operations required to turn one sequence to another. The distance considers insertions, deletionsm substitutions and transpositions.}
}
\newglossaryentry{DM}
{
        name=Data Mining,
        description={A process of extracting and discovering patterns in large data sets involving methods at the intersection of machine learning, statistics, and database systems.}
}
\newglossaryentry{bbm}
{
        name=blackbox model,
        description={A model, whose reasoning pattern is intransparent and incomprehensible for a human observer.}
}
\newdualentry{CSV}{CSV}{Comma Seperated Values}{A structured data format to store information. Every line relates to a data point and every feature is seperated by a seperator. The seperator is commonly a comma but other characters like tabs or semicolons are valid as well.}
\newdualentry{XES}{XES}{eXtensible Event Stream}{An XML-based data format to store event logs. The format was developed and adopted by the IEEE Task Force on Process Mining.}
\newdualentry{NLP}{NLP}{Natural Language Processing}{A discipline that is mainly concerned with the analysis and modelling of natural language.}
\newdualentry{GAN}{GAN}{Generative Adversarial Model}{A model in which two neural network train simultaneously. The generative model tries to generate instance examples, while the discriminative models tries to distinguish real examples from generated ones. Both, the discriminator and the generator can be used in isolation afterwards.}
\newdualentry{PM}{PM}{Process Mining}{A subdiscipline of Data Mining, which uses process logs to analyse and utilise data which was produced by processes.}
\newdualentry{MDP}{MDP}{Markov Decision Process}{A probablistic process which assumes that an agent can influence the outcome of the process by choosing decisions given a state and expecting a return for its performance.}
\newdualentry{MP}{MP}{Markov Process}{A probablistic process whose outcomes depend on the process' state.}
\newdualentry{RL}{RL}{Reinforcement Learning}{An are within Machine Learning, which seeks to allow an intelligent agent to choose the right actions by maximizing the cumulative rewards of a task setting.}
\newdualentry{SCM}{SCM}{Structural Causal Model}{Also called \emph{Structural Equation Model (SEM)}. An SCM is a set of variables and equations. The equations' the relationship of outputs from other equations and the variables. The relationship between functions and variables inform the causal relationship. SCMs can be represented as directed graphs in which nodes describe variables and equations and the vertices dependencies. Furthermore, a SCM can produce a dataset containing all the values thhat where generated.}
\newdualentry{HCI}{HCI}{Human Computer Interaction}{A cross-disciplinary discipline which observes and researches the interfaces and interactions between humans and computers.}

\newdualentry{IS}{IS}{Information System}{A technical system designed to collect, process, store, and distribute information.}
\newdualentry{BI}{BI}{Business Intelligence}{Discipline which focuses on gaining business related insights to information by the use of data analytics.}
\newdualentry{BPM}{BPM}{Business Process Management}{Discipline which uses various methods to discover, model, analyze, measure, improve, optimize, and automate business processes.}
\newacronym{TQM}{TQM}{Total Quality Management}
\newacronym{CPM}{CPM}{Corporate Performance Management}
\newacronym{CPI}{CPI}{Continuous Process Improvement}
\newacronym{XAI}{XAI}{Explainable AI}
\newacronym{ML}{ML}{Machine Learning}
\newacronym{VAE}{VAE}{Variational Autoencoder}
\newacronym{DMM}{DMM}{Deep Markov Model}
\newacronym{DKF}{DKF}{Deep Kalman Filter}
\newacronym{ELBO}{ELBO}{Evidence Lower-Bound}
\newacronym{HMM}{HMM}{Hidden Markov Model}
\newacronym{PGM}{PGM}{Probalistic Graphical Model}
\newacronym{BPMN}{BPMN}{Business Process Modell Notation}
\newacronym{LSTM}{LSTM}{Long Short-Term Memory}
\newacronym{RNN}{RNN}{Recurrent Neural Network}
\newacronym{CNN}{CNN}{Convolutional Neural Network}
\newacronym{NN}{NN}{Neural Network}
\newacronym{SSDLD}{SSDLD}{Semi-strucured Damerau-Levenshtein distance}
\newacronym{GA}{GA}{Genetic Algorithm}
\newacronym{GP}{GP}{Genetic Programming}
\newacronym{ES}{ES}{Evolutionary Strategy}
\newacronym{EP}{EP}{Evolutionary Programming}
\graphicspath{{figures/}}

\usepackage{subfiles} 

\begin{document}
\title{CREATED: Generating Viable Counterfactual Sequences for Predictive Process Analytics
% using an Evolutionary Algorithm for Event Data
% \title{CREATED: The Generation of viable Counterfactual Sequences using an Evolutionary Algorithm for Event Data of Complex Processes
% \thanks{Supported by organization x.}
}
\author{Olusanmi Hundogan\inst{1}\orcidID{0009-0001-5378-5388}\and
Xixi Lu\inst{1}\orcidID{0000-0002-9844-3330} \and
Yupei Du\inst{1} \and
Hajo A. Reijers\inst{1}\orcidID{0000-0001-9634-5852}}
\authorrunning{Hundogan et al.}
\titlerunning{CREATED}
\institute{Utrecht University, Utrecht, The Netherlands \\
\email{\{o.a.hundogan, x.lu, y.du, h.a.reijers\}@uu.nl} 
% \and
% TODO
% \and
}
\maketitle              % typeset the header of the contribution
\begin{abstract}
    % Within the field of Process Mining, deep recurrent networks (such as LSTM) have been used to predict the next state or the outcome of a multivariate sequence. However, these models tend to be complex and are difficult for users to understand of the underlying process model. Counterfactuals answer "what-if" questions, which are used to understand the reasoning behind the predicted outcome of a process. Current methods to generate counterfactual explanations do not take the structural characteristics of multivariate discrete sequences into account. In this work we propose a framework that uses evolutionary methods to generate counterfactuals, while incorporating criteria that ensure their viability. Our results show that it is possible to generate counterfactuals that are viable and automatically align with the factual. The generated counterfactuals outperform baseline methods in viability and yield comparable results compared to other methods in the literature.

Predictive process analytics %is an emerging research field in process mining that 
focuses on predicting future states, such as the outcome of running process instances. These techniques often use machine learning models or deep learning models (such as LSTM) to make such predictions. 
However, these deep models are complex and difficult for users to understand.
%
% \emph{Counterfactuals} are alternative execution  of a case that lead to a different outcome, help to understand the reasoning of the predictions of the deep models.
%
Counterfactuals answer ``what-if'' questions, which are used to understand the reasoning behind the predictions. For example, what if instead of emailing customers, customers are being called? Would this alternative lead to a different outcome? 
Current methods to generate counterfactual sequences either do not take the process behavior into account, leading to generating invalid or infeasible counterfactual process instances, or heavily rely on domain knowledge. 
In this work, we propose a general framework that uses evolutionary methods to generate counterfactual sequences. Our framework does not require domain knowledge. Instead, we propose to train a Markov model to compute the \emph{feasibility} of generated counterfactual sequences and adapt three other measures (\emph{delta} in outcome prediction, \emph{similarity}, and \emph{sparsity}) to ensure their overall viability. 
The evaluation shows that we generate viable counterfactual sequences, outperform baseline methods in viability, and yield similar results
% equally good results 
when compared to the state-of-the-art method that requires domain knowledge. 
% Counterfactual
% explainable AI
% predictive process analytics
% evolutionary algorithms
\end{abstract}
%
%
%
% TODO: Put all the thesis content in the paper
% TODO: Change chapters to sections
% NOTE: Technical papers describe original solutions (theoretical, methodological or conceptual) in the field of IS Engineering. A technical paper should clearly describe the situation or problem tackled, the relevant state of the art, the position or solution suggested and its potential‚ as well as demonstrate the benefits of the contribution through a rigorous evaluation.
% NOTE: Limit is 16 pages

% XIXI: Change cite_author to cite.
\section{Introduction}
\label{ch:intro}

%XL: storyline following the abstract
% Predictive process monitoring is ... and uses deep recurrent networks (such as LSTM) to predict next activity or process outcome. 
% The models tend to be complex and are difficult to explain. 
% 

% \subfile{content/sections/sec_motivation}

Predictive process analytics is an emerging research field in the process mining discipline that focuses on predicting the future states or outcome of running cases of business processes. The proposed techniques often use Machine Learning (ML) models or deep learning models (such as LSTM). These predictive models are trained on historical executions of business processes (i.e.,~\emph{event logs}) to make predictions of future states or outcomes. 
Studies have shown that predictive models can forecast the outcome of processes from various domains well~\cite{klimek_Longtermseriesforecasting_2021,tax_PredictiveBusinessProcess_2017}. For instance, in the medical domain, predictive models are applied to predict the outcome or trajectory of a patient's condition~\cite{mannhardt_Analyzingtrajectoriespatients_2017}. %In the private sector, predictive models are used to detect faults or outliers~\cite{}. 
% The research discipline \gls{DL} has shown promising results within domains that have been considered difficult for decades. The Moravex Paradox\cite{agrawal_studyphenomenonMoravec_2010}, which postulates that machines are capable of doing complex computations easily while failing in tasks that seem easy to humans such as object detection or language comprehension, does not hold anymore. Meaning that with enough data to learn, machines are capable of learning highly sophisticated tasks better than any human. The same holds for predictive tasks. 
% While many predictive models can predict certain outcomes, it remains a difficult challenge to understand their reasoning. 

While these predictive models are very powerful, they are usually complex and difficult to comprehend. Therefore, they are also known as \emph{blackbox models}. A lack of comprehension is undesirable for many application domains. For example, not knowing why a mortgage application was denied makes it impossible to rule out possible biases. In critical domains like medicine, the reasoning behind decisions becomes more crucial. For instance, if we know that a treatment process of a patient reduces the chances of survival, we want to know which treatment step is the critical factor we ought to avoid. For the engineering of fair and effective information systems, it is essential to comprehend and explain the reasoning behind predictions. 
% To summarise, knowing the outcome of a process often leads us to questions on how to change it. Formally, we would like to change the outcome of a process instance by making it maximally likely with as little interventions as possible~\cite{molnar2019}. 
% \autoref{fig:desired_outcome} is a visual representation of the desired goal.

% \begin{figure}[htb]
%     \centering
%     \includegraphics[width=0.99\textwidth]{figures/counterfactual_goal.png}
%     \caption{This figure illustrates a model, that predicts a certain trajectory of the process. However, we want to change the process steps in such a way, that it changes the outcome.}
%     \label{fig:desired_outcome}
% \end{figure}

% One way to better understand the \gls{ML} models lies within the \gls{XAI} discipline. XAI focuses the developments of theories, methods, and techniques that help explaining \glspl{bbm} to humans. Most of the discipline's techniques produce explanations that guide our understanding. Explanations can come in various forms, such as IF-THEN rules~\cite[p.90]{molnar2019} or feature importance~\cite[p.45]{molnar2019}.%, but some are more comprehensible for humans than others. 

% To understand the underlying reasoning of ML models, a human-friendly approach known as \emph{counterfactuals} is proposed within the eXplaibale AI (XAI) discipline~\cite[p. 221]{molnar2019}.
The \gls{XAI} discipline proposes \emph{counterfactuals} as a human-friendly approach to understanding the underlying reasoning of ML models~\cite[p. 221]{molnar2019}.
Counterfactuals can help us answer hypothetical ``what-if'' questions. In other words, assuming we know \emph{what} would happen \emph{if} we changed the execution of a process instance, we could change it for the better. For example, what if instead of emailing customers, customers are contacted by phone? Would this alternative sequence have led to a different outcome (e.g., instead of rejecting the offer, the customer accepts the offer)?

% In this paper, we raise the question how we can use counterfactuals to change the trajectory of a process models' prediction towards a desired outcome. Knowing the answers not only increases the understanding of \glspl{bbm}, but also help us avoid or enforce certain outcomes.
Existing methods can be divided into two categories: traditional and process-aware. The \emph{traditional} counterfactual methods focus on static, tabular data, such as DICE~\cite{mothilal_ExplainingMachineLearning_2020}. These methods aim to minimize the feature changes while maximize the flip in the outcome prediction. 
% For example, the methods may find that one change to the loan amount of a mortgage application would led to the predictive model flipping from an acceptance to a rejection. %\textcolor{red}{\textbf{TODO: maybe some challenges can be moved here.}}
%
% However, little research has focused on counterfactuals for dynamic data, in particular for event logs of business processes.
These methods do not take the process behavior into account. Applying them directly to event logs may lead to generating \emph{invalid} or \emph{infeasible} counterfactual sequences. 
% For example, a normative business process may specify that a loan application, if cancelled, then it cannot be accepted by the customer. Without taking the normative process behavior into account, the counterfactual methods may propose an invalid sequence, namely first cancelling a loan application, then immediately accept the offer.  
%
The \emph{process-aware} methods adapt the traditional methods for counterfactual generations of event logs~\cite{hsieh_DiCE4ELInterpretingProcess_2021}. While taking normative process behavior into account, these state-of-the-art methods, however, heavily rely on domain knowledge (e.g., users need to know the flows between milestones of a process)~\cite{hsieh_DiCE4ELInterpretingProcess_2021}.
In this paper, we approach the problem of generating counterfactual sequences for process outcome prediction without domain knowledge. In particular, we propose a general framework that uses evolutionary algorithms to generate sequences. The framework contains three components. The first component is a pre-trained predictive model, which we require to explain using counterfactuals. We assume that the prediction model \emph{accurately} predicts the outcome of a process at any step\footnote{\scriptsize The accuracy-condition is favorable, but not necessary. If the component is accurately modelling the real world, we can draw real-world conclusions from the explanations generated. If the component is inaccurate, the counterfactuals only explain the prediction decisions and not the real world.}. The second component implements the evolutionary algorithm, which generates counterfactual sequences that should be of high quality. To quantify the quality of counterfactual sequences and select the best ones, we define a \emph{viability} measure as our third component, which takes four measures into account, namely (1) feasibility of a counterfactual sequence, (2) the delta flipped in the outcome prediction, (3) the similarity between factual and counterfactual, and (4) the sparsity counting the number of changes. 
% To ensure the generated counterfactuals are viable and to rank and select the most viable counterfactuals, the third component implements the evaluation measures. 
%
As we use evolutionary algorithms to generate our counterfactuals, we refer to this framework as CREATED: the \textbf{C}ounte\textbf{R}factual Sequence generation with \textbf{E}volutionary \textbf{A}lgori\textbf{T}hms on \textbf{E}vent \textbf{D}ata. The name reflects how our model CREATEs new counterfactual sequences.

To evaluate the CREATED framework, we used ten event logs from three real-life processes and performed two experiments. First, we examined 54 configurations of the CREATED framework to obtain optimal configurations and compared our results with three baseline methods (case-based, sample-based, and random). The results show that we outperform the baseline methods in viability. In the second experiment, we compared our counterfactual sequences to the ones generated by a state-of-the-art method, showing that we yield similar counterfactuals without requiring domain knowledge.

% The evaluation shows that we generate viable counterfactual sequences, outperform baseline methods in viability, and yield similar results when compared to the state-of-the-art method that requires domain knowledge. 

% A plausible counterfactual is one whose outcome can be predicted by the predictive component. If the predictive component cannot predict the counterfactual sequence, we can assume that the generative model is \emph{unfaithful} to the predictive component that we want to explain. The third component is the evaluation metric upon which we decide the viability of the counterfactual candidates.

% We implemented our framework as follows. We used a LSTM model for the predictive component. For the generative component, we propose an evolutionary algorithm, while

% For the evaluation, we have to show the following:
% \begin{itemize}
%     \item[RQ2-H1:] If we use a viability function which incorporates multiple criteria to determine counterfactuals, we consistently retrieve more viable counterfactuals, than choosing the counterfactuals the at random.
%     \item[RQ2-H2:] The generated counterfactuals consistently outperform the most viable counterfactuals among examples in the dataset.
%     \item[RQ3-H1:] The results of the counterfactual are comparable to other existing literature.
%           % \item[RQ2-H1:] The counterfactual generation consistently identifies the most viable counterfactual in the dataset faster than a random search.
% \end{itemize}

% \subsection{Outline}
% \subfile{content/sections/sec_outline}
The remainder of the paper is structured as follows.
Sections~\ref{ch:relatedwork} and~\ref{sec:formulas} respectively discuss the related work and preliminary concepts.
Section~\ref{ch:methods} presents our approach. Section~\ref{ch:evaluation} explains the evaluation set-up. 
Section~\ref{ch:results} discusses the results, and Section~\ref{ch:conclusion} concludes the paper. 

% In \autoref{ch:prereq}, we introduce all of the important concepts that are crucial to this thesis. Most importantly, we introduce the main research discipline \Gls{PM} and the subject of our research: \emph{Counterfactuals}. Furthermore we cover some necessary background required to understand the methods, we employ.
% The \autoref{ch:methods}, introduces our methodological framework in further detail. The chapter explains all the important components and methods, we apply, to answer the research question. Among these methods, we introduce the methodological architecture, a modified version of the \Gls{damerau_levenshtein}.
% \autoref{ch:evaluation} covers the main approach behind our experimental setup. We discuss how we attempt to answer our research questions and introduce the datasets we are using and how we conduct the preprocessing. 
% In \autoref{ch:results} we report on the results and insights we gain from executing our research approach. 
% All the results are summarised in \autoref{ch:discussion}. Here, we summarize and interpret our results. We discuss limitations and possible improvements. We also discuss implications for future research endeavors. 
% The \autoref{ch:conclusion} summarizes the thesis and the implications for the \Gls{PM} research field.

\section{Related work}
\label{ch:relatedwork}
% % \subsection{Related Literature}
% \label{sec:literature}

% Many researchers have worked on counterfactuals and \Gls{PM}. 
% Here, we combine the important concepts and discuss the various contributions to this paper.
% \subfile{content/sections/sec_literature}

% \textbf{Generating Counterfactuals}
As stated before, We divide the existing methods for counterfactual generation into two categories: \emph{traditional} methods and \emph{process-aware} methods. The \emph{traditional} methods concern the classical ML models, and the topic of counterfactual generation as an explanation method was first introduced by \cite{wachter_CounterfactualExplanationsOpening_2017}. The authors defined a loss function that incorporates the criteria to generate a counterfactual that maximizes the likelihood of a predefined outcome and minimizes the distance to the original instance.
% However, the solution of~\cite{wachter_CounterfactualExplanationsOpening_2017} did not account for the minimisation of feature changes and does not penalize unrealistic features. Furthermore, their solution cannot incorporate categorical variables.
%
A more recent approach by~\cite{dandl_MultiObjectiveCounterfactualExplanations_2020} incorporates four main criteria for counterfactuals by applying a genetic algorithm with a multi-objective fitness function~\cite{dandl_MultiObjectiveCounterfactualExplanations_2020}. This approach strongly differs from gradient-based methods, as it does not require a differentiable objective function. However, the above traditional methods focus on static data. They do not take process behaviors into account. Applying these methods directly on event logs may result in generating infeasible counterfactual sequences. 

Within process mining, the \emph{process-aware} methods for counterfactuals have followed two streams. The first steam uses the \gls{causalinference} techniques to analyse and model business processes, as the causal relationships can be used to understand the effect of decisions in a process on its outcome. However, early work has often attempted to incorporate domain-knowledge about the causality of processes in order to improve the process model itself~\cite{baker_ClosingLoopEmpirical_2017,hompes_DiscoveringCausalFactors_2017,shook_assessmentusestructural_2004,wang_CounterfactualDataAugmentedSequential_2021}.
Among these, the approach in~\cite{narendra_CounterfactualReasoningProcess_2019} is one of the first to include counterfactual reasoning for process optimization~\cite{narendra_CounterfactualReasoningProcess_2019}.
Later, the work by~\cite{oberst_CounterfactualOffPolicyEvaluation_2019} uses counterfactuals to generate alternative solutions to treatments, which lead to a desired outcome.
However, the authors do not attempt to provide an explanation of the model's outcome and therefore, disregard multiple viability criteria for counterfactuals in \gls{XAI}. 
\cite{qafari_CaseLevelCounterfactual_2021} published the most recent paper on the counterfactual generation of explanations. The authors use a known \gls{SCM} to guide the generation of their counterfactuals. However, this approach requires a process model which is as close as possible to the \emph{true} process model. Our approach assumes no knowledge of such a normative process model.

% Within the \gls{XAI} context, \cite{tsirtsis_CounterfactualExplanationsSequential_2021} developed the first explanation method for process data. However, their work closely resembles the work of \cite{oberst_CounterfactualOffPolicyEvaluation_2019} and treat the task as \gls{MDP}\cite{oberst_CounterfactualOffPolicyEvaluation_2019}. This extension of a regular \gls{MP} assumes that an actor influences the outcome of a process given the state. This formalisation allows the use of \gls{RL} methods like Q-learning or SARSA. However, this often requires additional assumptions such as a given reward function and an action-space. For counterfactual sequence generation, there is no obvious choice for the reward function or the action-space. 
% Nonetheless, both \cite{tsirtsis_CounterfactualExplanationsSequential_2021} and \cite{oberst_CounterfactualOffPolicyEvaluation_2019} contribute an important idea - the idea of incrementally generating the counterfactual instead of the full sequence. 

The second stream in \emph{process-aware} methods adapts the \emph{traditional} counterfactual methods for process-aware counterfactuals. The DICE4EL approach~\cite{hsieh_DiCE4ELInterpretingProcess_2021} extends the DICE method~\cite{mothilal_ExplainingMachineLearning_2020} to generate counterfactuals for event logs while building on the same notion of incremental generation. 
%
% Their approach has a very similar structure to our approach and appears to be the only one that we can compare our counterfactuals against. 
% For this reason, this paper highlights some key differences and similarities. However, to understand the differences and similarities, we first have to establish some core concepts.  In this section, we only discuss their approach, briefly.
%
The authors recognised that some processes have critical events (mile-stones) which govern the overall outcome. Hence, by simply avoiding the undesired outcome from critical event to critical event, it is possible to limit the search space and compute viable counterfactuals. However, their approach requires concrete domain knowledge about these critical points. We propose a framework that avoids this constraint and does not require domain knowledge. The LORELEY approach \cite{DBLP:conf/emcis/HuangMP21} extends the LORE method~\cite{DBLP:journals/expert/GuidottiMGPRT19} and also uses an evolutionary algorithm. However, this approach focuses on mutating the case/event attributes. More specifically, the approach treats the encoded features representing the control flow as a single attribute in the crossover and mutation steps; thus, no unseen counterfactual sequences are created. In contrast, we generate unseen process sequences. Furthermore, we propose to automatically train a Markov model from the input event log to capture the likelihood of a process sequence. This Markov model is then used to derive the feasibility of counterfactual sequences. 

\section{Background}
% \subsection{Formal Definitions}
\label{sec:formulas}
% Before diving into the rest of this thesis, we have to establish preliminary definitions, we use in this work. With this definitions, we share a common formal understanding of mathematical descriptions of every concept used within this thesis. 

% XIXI: Make much shorter like the example but add your new concepts
% \subsection{Process Logs, Cases and Instance Sequences}
We start by formalising the event log and its elements.

\noindent\textbf{Definition 1: Case, Event and Log} 
Let $\mathcal{E}$ be the universe of the event identifiers and $E \subseteq \mathcal{E}$ a set of events. An event log $L \subseteq \mathcal{E}^*$ is a set of sequences of events. 
Let $C$ be a set of case identifiers and $\pi_c : E \mapsto C$ a surjective function that links every element in $E$ to a case $c \in C$ in which $c$ signifies a specific case. 
For a set of events $E \subseteq \mathcal{E}$, the shorthand $s^c$ denotes a particular sequence $s^c = \langle e_1, e_2, \ldots, e_t \rangle$ with $c$ as case identifier and a length of $t$. Each $s$ is a trace of the process log $s \in L$.  
Let $\mathcal{T}$ be the time domain and $\pi_t : E \mapsto \mathcal{T}$ a non-surjective linking function which strictly orders a set of events. 
% Let $\mathcal{A}$ be a universe of attribute identifiers, in which each identifier maps to a set of attribute values $\overline{a}_i \in \mathcal{A}$. 
% Let $\overline{a}_i$ correspond to a set of possible attribute values by using a surjective mapping function $\pi_A : \mathcal{A} \mapsto A$. 
Each event $e_t$ consists of a set $e_t = \{ a_1 \in A_1, a_2 \in A_2, \ldots, a_I \in A_I\}$ with the size $I = |A|$, in which $A_i$ is an attribute and $a_i$ represents a possible value of that attribute. 
% We define a mapping from an attribute value to its respective attribute identifier $\pi_{\overline{a}} : A \mapsto \mathcal{A}$. Hence, we can map every event attribute value back to its attribute identifier. 

% \subsection{Representation}
\noindent\textbf{Definition 2: Attribute Representation}
Let $\pi_d : A_i \mapsto \mathbb{N}$ be a surjective function, which determines the dimensionality of $a_i$, and let $F$ be a set of size $I$ containing a representation function for every attribute. Let $f_i \in F$ be mapping functions to a vector space $f_i : a_i \mapsto \mathbb{R}^d_i$, in which $d$ represents the dimensionality of an attribute value $d = \pi_d(A_i)$. 
We denote any event $e_t \in s^c$ of a specific case $c$ as a vector, which concatenates every attribute representation $f_i$ as $\mathbf{e}_t^{c} = [f_1; f_2; \ldots; f_I]$. Therefore, $\mathbf{e}_t^{c}$ is embedded in a vector space of size $D$ which is the sum of each individual attribute dimension $D = \sum_i \pi_d(A_i)$. In other words, we concatenate all representations, whether they are scalars or vectors to one final vector representing the event. Furthermore, if we refer to a specific attribute $A_i$, we use the shorthand $\overline{a}_i$. 

% \autoref{fig:representation} shows a schematic representation of a log $L$, a case $c$ and an event $e$.

% \begin{figure}[htbp]
%     \centering
%     \includegraphics[width=0.9\textwidth]{figures/Graphics/Slide4.PNG}
%     \caption{This figure shows the representation of a log $L$ which contains a number of cases $s$. Case $s^2$ contains a number of events $e_t$. Each events has attribute values $a_i$, which are mapped to vector spaces of varying dimensions. At last, all of the vectors are concatenated.}
%     \label{fig:representation}
% \end{figure}

% REMOVED state space models formal description

% \subsection{Representation}
% \label{sec:representation}
% \subfile{content/sections/sec_representation}

\section{Methods}
\label{ch:methods}
% In this chapter, we describe details of our framework and discuss advantages and limitations. 
% Therefore, we provide a more detailed overview and additionally describe all components. As the framework resembles the work of \cite{hsieh_DiCE4ELInterpretingProcess}, we also discuss differences and similarities between both solutions. 

\subsection{Methodological Framework: CREATED}
\label{sec:framework}
\begin{figure}[htb]
    \centering
    \includegraphics[width=0.99\textwidth]{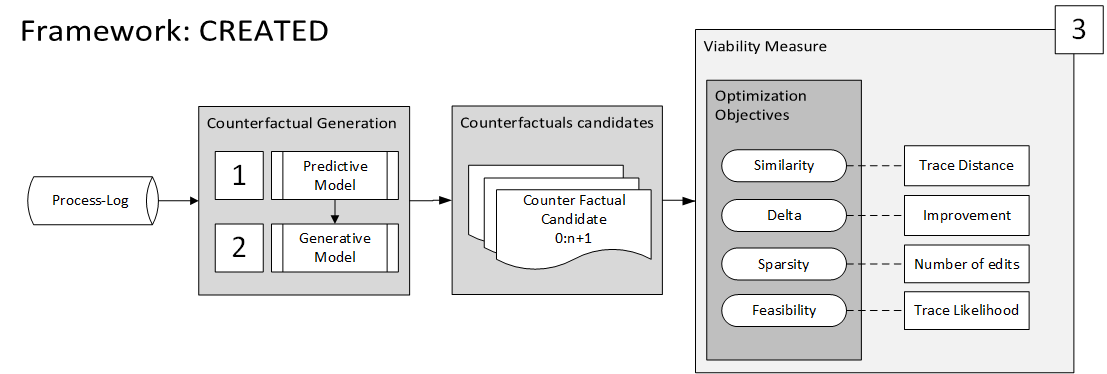}
    \caption{The CREATED framework: the input is the process log; the log is used to train a predictive model (Component 1) and the generative model (Component 2). This process produces a set of candidates which are subject to evaluation via the validity metric (Component 3).}
    \label{fig:approach}
\end{figure}

% \subsubsection{Architecture}
% \subfile{content/sections/sec_framework}
To generate counterfactuals, we need to establish a conceptual framework consisting of three main components. The three components are shown in \autoref{fig:approach}. 

% \attention{Change the names of the measures.}

The first component is a \emph{predictive model}. As we attempt to explain model decisions with counterfactuals, the predictive model needs to be pretrained. We can use any model that can predict the probability of a sequence. The prediction model in this paper is a simple LSTM model using the process log as an input. The architecture is inspired by \cite{hsieh_DiCE4ELInterpretingProcess_2021}. The model is trained to predict the outcome given a sequence. 

The second component is a \emph{generative model}. The generative model produces counterfactuals given a factual sequence. We implement an evolutionary generator that takes a factual as input and yields counterfactuals candidates as output.
% Specifically, we compare an evolutionary counterfactual generator against three naive generative baseline approaches. These baselines do not optimise towards the viability criteria.  All approaches allow us to use a factual sequence as a starting point for the generative production of counterfactuals. Furthermore, they also generate multiple counterfactual candidates. 

The generated candidates are subject to the third major component. 
To select the most \emph{viable} counterfactual candidate, we evaluate their viability score using a custom metric. 
The metric incorporates four criteria for viable counterfactuals. 
We measure the \textbf{similarity} between two sequences using a multivariate sequence distance metric. The \textbf{outcome-delta} is the difference between the likelihood of the factual and the counterfactual. For this purpose, we require the predictive model, which computes a prediction score reflecting the likelihood. 
We measure \textbf{sparsity} by counting the number of changes in the features and computing the edit distance. Lastly, we need to determine the \textbf{feasibility} of a counterfactual. We measure the feasibility by estimating the probability of a counterfactual. %
Note that our method was developed for outcome prediction but can be adapted to the next activity prediction task.

\subsection{Counterfactual Generators}
\label{sec:model_generation}
\textbf{Generative Model: Evolutionary Algorithm}
\label{sec:model_evolutionary}
% \subfile{content/sections/sec_model_evolutionary}
% We introduced most operator types in \autoref{sec:evo}.
In this section, we describe the concrete set of operators and select a subset that we want to explore further.

For our purposes, the \emph{gene} of a sequence consists of the sequence of events within a \gls{instance}. Hence, if an offspring inherits one parent gene, it inherits the activity associated with the event and its event attributes. Our goal is to generate candidates by evaluating the sequence based on our viability measure. Our measure acts as the fitness function. The candidates that are deemed fit enough are subsequently selected to reproduce offspring. This process is explained in \autoref{fig:example-inheritance}. 

\begin{figure}[t]
    \begin{tikzpicture}[>=stealth,thick,baseline]
        \matrix [matrix of math nodes, left delimiter=(,right delimiter=)](m1){
            a   & b    & a    \\
            0.6 & 0.25 & 0.70 \\
            0   & 0    & 1    \\
            1.2 & 4.5  & 2.3  \\
        };
        \node[draw,  blue!80, dashed, thin, inner sep=2mm,fit=(m1-1-1.west) (m1-4-2.east)] (attbox) {};
        \node[above = 2mm of m1](rlbl) {Parent 1};
        \node[below = 2mm of attbox, blue!80](lb1) {\tiny Genes passed on};

        \node[right = 2mm of m1](cr){+};

        \matrix [matrix of math nodes,left delimiter=(,right delimiter=), right = 2mm of cr](m2){
            a   & b    & a    & c    \\
            0.6 & 0.75 & 0.64 & 0.57 \\
            0   & 0    & 1    & 0    \\
            1.2 & 4.5  & 3.3  & 3.0  \\
        };
        \node[draw,  red!80, dashed, thin, inner sep=2mm,fit=(m2-1-3.west) (m2-4-4.east)] (attbox-m2) {};
        \node[above = 2mm of m2](rlbl) {Parent 2};
        \node[below = 2mm of attbox-m2, red!80](lb2) {\tiny Genes passed on};

        \node[right = 2mm of m2](eq){=};

        \matrix [matrix of math nodes,left delimiter=(,right delimiter=), right = 2mm of eq](m3){
            a   & b    & a    & c    \\
            0.6 & 0.25 & 0.64 & 0.57 \\
            0   & 0    & 1    & 0    \\
            1.2 & 4.5  & 3.3  & 3.0  \\
        };

        \node[draw,  blue!80, dashed, thin, inner sep=2mm,fit=(m3-1-1.west) (m3-4-2.east)] (attbox-m3-1) {};
        \node[draw,  red!80, dashed, thin, inner sep=2mm,fit=(m3-1-3.west) (m3-4-4.east)] (attbox-m3-2) {};
        \node[below = 2mm of attbox-m3-1, blue!80](lb2) {\tiny Inherited};
        \node[below = 2mm of attbox-m3-2, red!80](lb2) {\tiny Inherited};

        \node[above = 2mm of m3](rlbl) {Offspring};
        % \draw[->, outer] (m2.east) -- (m3.west);
    \end{tikzpicture}
    \caption{A newly generated offspring inheriting genes in the form of activities and event attributes from both parents.}
    \label{fig:example-inheritance}
\end{figure}

The offspring is subject to mutations. We evaluate the new population and repeat the procedure until a termination condition is reached. We can optimise the viability measure established in Section~\ref{sec:viability}.

\newcommand{\cf}{\text{counterfactuals}}
\newcommand{\cfp}{\text{cf-parents}}
\newcommand{\cfo}{\text{cf-offsprings}}
\newcommand{\cfm}{\text{cf-mutants}}
\newcommand{\cfs}{\text{cf-survivors}}

\begin{algorithm}[b]
    \caption{The basic structure of an evolutionary algorithm.}
    \begin{algorithmic}
        \Require{factual, configuration, sample-size, population-size, mutation-rate, termination-point}
        % \Require{}
        % \Require{}
        % \Require{}
        % \Require{}
        % \Require{}
        \Ensure{The result is the final counterfactual sequences}

        \State $counterfactuals \gets initialize(\text{factual})$
        \While{not $termination$}
        \State $\cfp \gets select(\cf, \text{sample-size})$
        \State $\cfo \gets crossover(\cfp) $
        \State $\cfm \gets mutate(\cfo, \text{mutation-rate})$
        \State $\cfs \gets recombine(\cf, \cfm, \text{population-size})$
        \State $termination \gets determine(\cfs, \text{termination-point})$
        \State $\cf \gets \cfs$
        \EndWhile
    \end{algorithmic}
    \label{alg:my-evolutionary}
\end{algorithm}

\noindent\textbf{Operators}
We implemented several different evolutionary operators. Each one belongs to one of five categories. The categories are initiation, selection, crossing, mutation, and recombination. Table~\ref{tab:abbreviation} contains a complete list of the operators.

% XIXI: Shorten many of the descriptions by describing a standard reference book.
% XIXI: Likes the images for OPC, TPC and UPC

\begin{table}
\begin{center}
\caption{An overview of all evolutionary operators used in this paper and a short description.}
\label{tab:abbreviation}
% \begin{small}
\begin{tabular}{ | m{0.8cm} | m{2.5cm}| m{8.5cm} | } 
% \begin{tabular}{ |l|l|l| } 
\hline
    Label & Name & Description\\
    
    \hline
    % \hline
    \multicolumn{3}{|l|}{Initiation}\\
    \hline
    RI & Random Initialisation & Generates an initial population in which the event sequence was chosen at random based on the log. The event attributes were drawn from a normal distribution. \\ 
    SBI & Sampling-Based Initialisation & Generates an initial population by sampling from a data distribution estimated from the data directly. The event sequence was sampled using the event transition probabilities. The attributes were sampled using distributions conditioned on the emitted events.\\ 
    CBI & Case-Based Initialisation & Samples initial population directly from the Log. \\ 
    \hline
    % \hline
    \multicolumn{3}{|l|}{Selection}\\
    \hline
    RWS & Roulette-Wheel-Selection & Selects individuals randomly in proportion to their fitness value  \\
    TS & Tournament-Selection & Selects pairs of individuals and compares each pair. The better individual between both pairs has a higher chance of being selected.\\
    ES & Elitism-Selection & Selects individual with the highest fitness. \\
    \hline
    % \hline
    \multicolumn{3}{|l|}{Crossover}\\
    \hline    
    UCx & Uniform Crossover & uniformly choose a fraction of genes of one individual (\emph{Parent 1}) and overwrite the respective genes of another individual (\emph{Parent 2}).\\
    OPC & One-Point Crossover & Chooses a point in the sequence and overwrites the genes of \emph{Parent 2} by the genes \emph{Parent 2} from that point onward.\\
    TPC & Two-Point Crossover & Chooses two points in the sequence and overwrites the sequence in between the two points from \emph{Parent 2} with the sequence from \emph{Parent 1}.\\
    \hline
    % \hline
    \multicolumn{3}{|l|}{Mutation}\\

    \hline    
    RM & Random-Mutation & Inserts, changes or deletes activities randomly. Event attributes are drawn from a normal distribution.\\
    SBM & Sampling-Based Mutation & Inserts, changes or deletes activities randomly. Event attributes are drawn from an estimated data distribution.\\
    \hline
    % \hline
    \multicolumn{3}{|l|}{Recombination}\\
    \hline   
    FSR & Fittest-Survivor Recombination & Strictly determines the survivors among the mutated offsprings and the current population by sorting them in terms of viability\\
    BBR & Best-of-Breed Recombination & Determines offsprings that are better than the average within their generation and adds them to survivors of past generations.\\
    RR & Ranked Recombination & selects the new population differently than the former recombination operators. Instead of using the viability directly, we sort each individuum by every viability component separately. This approach allows us to select individuals regardless of the scales of every individual viability measure.\\
    \hline
\end{tabular}
% \end{small}
\end{center}
\end{table}

\noindent\textbf{Naming-Conventions}
We use abbreviations to refer to each model configuration. For instance, \emph{CBI-RWS-OPC-RM-RR} refers to an evolutionary operator configuration that samples its initial population from the data (CBI), probabilistically samples parents based on their fitness (RWS), crosses them on one point (OPC), and so on. For the \emph{Uniform-Crossing} (UCx) operator, we additionally indicate its crossing rate using a number. For instance, \emph{CBI-RWS-UC3-RM-RR} uses the \emph{Uniform-Crossing} (UC3) operator. The child receives roughly 30\% of the genome of one parent and 70\% of another parent. 

\noindent\textbf{Hyperparameters}
The evolutionary approach comes with a number of hyperparameters. 
We first discuss the \emph{model configuration}. As shown in this section, there are a \NumEvoCombinations ways to combine all operators. Depending on each operator combination, we might see very different behaviours. 
The decision of the appropriate set of operators is by far the most important in terms of convergence speed and result quality.
The next hyperparameter is the \emph{termination point} which determines the duration of the search. 
Optimally, we find a termination point, which is not too early but not too late, too.
The \emph{mutation rate} is another hyperparameter. It signifies how much a child can differ from its parent.

\subsection{Viability Measure}
\label{sec:viability}

\subsubsection{Feasibility-Measure}
\label{sec:feasibility}
% \subfile{content/sections/sec_viability_feasibility}
To determine the feasibility of a counterfactual trace, it is important to consider two aspects. 
First, we have to compute the probability of the sequence of event transitions. This is a difficult task, given the \emph{Open World assumption}\footnote{In theory, we cannot know whether or not any event \emph{can} follow after another event.}. 
Therefore, we have to assume the data is representative and the underlying process is static. This assumption allows us to estimate first-order transition probabilities by counting event transitions.

% However, if the data is representative of the process dynamics, we can make simplifying assumptions. 
% For instance, we can compute the first-order transition probability by counting each transition. The issue remains that longer sequences tend to have a zero probability if they have never been seen in the data. 

Second, we have to compute the feasibility of the individual feature values given the sequence. We can relax the computation of this probability using the \emph{Markov Assumption}. In other words, we assume that each event vector depends on the current activity but on none of the previous events and features. This means that we can model density estimators for every event and use them to determine the likelihood of a set of features.

We define the feasibility measure in \autoref{eq:feasibility_measure}, where $e_t$ represents the current event, transited from the previous event $e_{t-1}$. Likewise, $f$ represents the emission of the feature attributes. Hence, the probability of a particular sequence is the product of the transition probability multiplied by the state emission probability for each step. 
% Note that this is the same as the feasibility measure in \autoref{eq:feasibility_measure}. 

\begin{align}
    \prob{e_{0:T},f_{0:T}} & = \prob{e_0}\cprob{f_0}{e_0}\prod_1^T \cprob{e_t}{e_{t-1}} \cprob{f_t}{e_t}
    \label{eq:feasibility_measure}
\end{align}

\subsubsection{Delta-Outcome}
\label{sec:delta}
% \subfile{content/sections/sec_viability_delta}
For the delta measure, we evaluate the likelihood of a counterfactual trace by determining whether a counterfactual leads to the desired outcome or not. For this purpose, we use the predictive model, which returns a prediction for each counterfactual sequence. As we are predicting process outcomes, we typically predict a class. However, forcing a deterministic model to produce a different class prediction is often difficult. Therefore, we can relax the condition by maximising the prediction score of the desired counterfactual outcome~\cite{molnar2019}. If we compare the difference between the counterfactual prediction score with the factual prediction score, we can determine an increase or decrease. Ideally, we want to increase the likelihood of the desired outcome. We refer to this value as \emph{delta}. For the binary outcome prediction case, we define the function as shown in \autoref{eq:delta}.

\begin{align}
    \label{eq:delta}
%     d_{a, b}(i, j) & =\min
    delta &= 
    \begin{cases}
            |p(o|s^*)-p(o|s)| &  \text{if }  p(o|s) > 0.5 \text { \& }  p(o|s) > p(o|s^*) \\                 
            -|p(o|s^*)-p(o|s)| &  \text{if }  p(o|s) > 0.5 \text { \& }  p(o|s) \leq p(o|s^*) \\                 
            |p(o|s^*)-p(o|s)| &  \text{if }  p(o|s) \leq 0.5 \text { \& }  p(o|s) > p(o|s^*) \\                 
            -|p(o|s^*)-p(o|s)| &  \text{if }  p(o|s) \leq 0.5 \text { \& }  p(o|s) \leq p(o|s^*) \\                 
    \end{cases} 
\end{align}

\subsubsection{Similarity Measure}
\label{sec:similarity}
We use a function to compute the \textbf{similarity} between the factual sequence and the counterfactual candidates. To incorporate differences in length between both sequences, we use a weighted version of the Damerau-Levenshtein distance~\cite{damerau_TechniqueComputerDetection_1964}. %
The Damerau-Levenstein distance applies a cost constant of 1 for each sequential difference. However, as process instances differ not only in event sequences but also in their event attribute values, we use a distance function to weigh the cost. In the case of \textbf{similarity}, we apply the euclidian distance. For formal definitions, we refer to~\cite[p.~42]{Hundogan2022}. 

% \vspace{-2em}
\subsubsection{Sparsity Measure}
\label{sec:sparsity}
For measuring the sparsity, we use the same weighted version of the Damerau-Levenshtein distance. However, to measure the distance, we count the number of differences between event attributes. For formal definitions, we refer to~\cite[p.~42]{Hundogan2022}.

% \textcolor{gray}{Furthermore, instead of defining the cost only in terms of event sequences, we incorporate cost function which takes the attribute values of an event into account. For measuring the similarity we rely on the euclidean distance.
% To measure the sparsity we simply count the differences between the event attributes. We typically want to minimize the number of changes. As with similarity, we incorporate structural differences between the two sequences by using the modified Damerau-Levenshtein distance.}

% \vspace{-2em}
\subsubsection{Viability-Measure}
\label{sec:viability}
We combine the feasibility measure, the outcome delta, the normalised sparsity, and normalised similarity measure by summation. As each measure can have values between 0 and 1, the viability measure ranges between 0 and 4. For more details on the viability measure, we refer to \cite[Chap.~3.3]{Hundogan2022}.

\section{Evaluation}
\label{ch:evaluation}
% In this section, we discuss the datasets, the preprocessing pipeline, and the final representation for each of the algorithms.  
% There, you will find instructions on how to install and run the experiments yourself.

\subsection{Datasets}
\label{sec:dataset_description}
% \subfile{content/sections/sec_dataset_stats}
% XIXI: Reference the dataset for outcome prediction and say we used this publically available benchmark datasets
For our evaluation, we use ten event logs of three real-life processes, which were also used in~\cite{teinemaa_OutcomeOrientedPredictiveProcess_2019}. Each dataset consists of events and contains labels that signify a process instance's outcome. We focus on binary outcome predictions. 
We include a variation of the BPIC dataset. This dataset was used in \cite{hsieh_DiCE4ELInterpretingProcess_2021}. The difference between Hsieh et al.'s dataset and the original dataset is two-fold. First, the authors focus on the generation of two event attributes. Second, the dataset is primarily designed for next-activity prediction, not outcome prediction. We modified the dataset to fit the outcome prediction model.
For more information about these datasets we refer to the comparative study by \cite{teinemaa_OutcomeOrientedPredictiveProcess_2019}. We list the important descriptive statistics in \autoref{tbl:dataset-stats}.

\begin{table}[t]
    \caption{All datasets used within the evaluation. DiCE4EL is used for the qualitative evaluation, and the remaining are used for quantitative evaluation purposes.}
    \label{tbl:dataset-stats}
    % \begin{adjustbox}{center}
        \makebox[\linewidth]{
            \begin{tabular}{lrrrrrrrrr}
\toprule
 & \#Cases & Min Len & Max Len & \% Unique Traces & \#Unique Ev. & \#Data Columns & \#Event Attr & \#Regular & \#Deviant \\
Dataset &  &  &  &  &  &  &  &  &  \\
\midrule
DiCE4EL & 3 051 & 12 & 25 & 0.000328 & 23 & 9 & 7 & 1 853 & 1 198 \\
BPIC12-25 & 3 051 & 12 & 25 & 0.000328 & 23 & 23 & 21 & 1 853 & 1 198 \\
BPIC12-50 & 4 587 & 12 & 50 & 0.000218 & 23 & 23 & 21 & 2 405 & 2 182 \\
BPIC12-75 & 4 677 & 12 & 75 & 0.000214 & 23 & 23 & 21 & 2 436 & 2 241 \\
BPIC12-100 & 4 685 & 12 & 96 & 0.000213 & 23 & 23 & 21 & 2 442 & 2 243 \\
Sepsis-25 & 707 & 5 & 25 & 0.001414 & 15 & 75 & 73 & 610 & 97 \\
Sepsis-50 & 770 & 5 & 47 & 0.001299 & 15 & 76 & 74 & 662 & 108 \\
Sepsis-75 & 777 & 5 & 66 & 0.001287 & 15 & 76 & 74 & 667 & 110 \\
Sepsis-100 & 779 & 5 & 88 & 0.001284 & 15 & 76 & 74 & 669 & 110 \\
TrafficFines & 129 615 & 2 & 20 & 0.000008 & 10 & 40 & 38 & 70 602 & 59 013 \\
\bottomrule
\end{tabular}

            }
            % \end{adjustbox}
\end{table}

% \subfile{content/sections/sec_dataset_preds}
\begin{table}[b]
    \caption{The evaluation metrics for the prediction component on all datasets. Includes precision, recall and f1 score for test, training and validation data.}
    \label{tbl:dataset-preds}
    % \begin{adjustbox}{center}
        \makebox[\linewidth]{
            \begin{tabular}{lrrrrrrrrrrrr}
\toprule
 & \multicolumn{3}{l}{precision} & \multicolumn{3}{l}{recall} & \multicolumn{3}{l}{f1-score} & \multicolumn{3}{l}{support} \\
Subset & test & training & validation & test & training & validation & test & training & validation & test & training & validation \\
Dataset &  &  &  &  &  &  &  &  &  &  &  &  \\
\midrule
BPIC12-100 & 1.000 & 0.999 & 0.999 & 1.000 & 0.999 & 0.999 & 1.000 & 0.999 & 0.999 & 60.000 & 1000.000 & 841.000 \\
BPIC12-25 & 0.808 & 0.770 & 0.765 & 0.750 & 0.742 & 0.733 & 0.738 & 0.733 & 0.723 & 60.000 & 1000.000 & 1000.000 \\
BPIC12-50 & 1.000 & 1.000 & 1.000 & 1.000 & 1.000 & 1.000 & 1.000 & 1.000 & 1.000 & 60.000 & 1000.000 & 819.000 \\
BPIC12-75 & 1.000 & 1.000 & 1.000 & 1.000 & 1.000 & 1.000 & 1.000 & 1.000 & 1.000 & 60.000 & 1000.000 & 841.000 \\
DiCE4EL & 0.780 & 0.806 & 0.821 & 0.700 & 0.755 & 0.749 & 0.677 & 0.744 & 0.739 & 60.000 & 1000.000 & 1000.000 \\
Sepsis-100 & 0.259 & 0.246 & 0.250 & 0.509 & 0.496 & 0.500 & 0.343 & 0.329 & 0.333 & 55.000 & 123.000 & 42.000 \\
Sepsis-25 & 0.478 & 0.511 & 0.528 & 0.483 & 0.508 & 0.519 & 0.449 & 0.482 & 0.495 & 60.000 & 1000.000 & 873.000 \\
Sepsis-50 & 0.250 & 0.240 & 0.261 & 0.500 & 0.490 & 0.511 & 0.333 & 0.322 & 0.346 & 60.000 & 1000.000 & 1000.000 \\
Sepsis-75 & 0.207 & 0.254 & 0.300 & 0.455 & 0.504 & 0.548 & 0.284 & 0.338 & 0.388 & 55.000 & 123.000 & 42.000 \\
TrafficFines & 1.000 & 0.987 & 0.984 & 1.000 & 0.987 & 0.983 & 1.000 & 0.987 & 0.983 & 60.000 & 1000.000 & 1000.000 \\
\bottomrule
\end{tabular}

            }
            % \end{adjustbox}
\end{table}

We list the predictions of our prediction component in \autoref{tbl:dataset-preds}. The F1-Scores on the test sets are generally higher for the BPIC dataset. Furthermore, in the case of the BPIC datasets, the prediction model always predicts the correct outcome if the max-length of the sequence exceeds 25. It is fair to assume that the length of a loan application process determines the chance of getting rejected or not.

\subsection{Preprocessing}
\label{sec:preprocessing}
% XIXI: The issue of using maximum 25 seq-len is a limitation that should be discussed again.
To prepare the data for our experiments, we employed basic tactics for preprocessing. First, we split the log into a training and a test set. 
Then, we filter out every case whose sequence length exceeds 25. We keep this maximum threshold for most experiments focusing on the evolutionary algorithm. The reason is the polynomial computation time of the viability measure. The similarity and sparsity components of the proposed viability measure have a runtime complexity of at least $N^2$. Hence, limiting the sequence length saves a substantial amount of temporal resources.
Next, we extract time variables if they are provided in the log. Then, we normalise the values. 
Each categorical variable is converted using binary encoding. 
The activity is label-encoded. As a result, every category is assigned to a unique integer. The label column is binary encoded, as we focus on outcome prediction.
Lastly, we pad each sequence towards the longest sequence in the dataset.

\subsection{Baseline Models}
We use three baseline models and compare them to the evolutionary models. The first baseline generates a random sequence of events and event attributes. Hence, we refer to this approach as \textbf{Random baseline} (RGW). We expect most models to perform better than this baseline. Otherwise, it would indicate that a random search would generate better counterfactuals than a guided one. The second baseline resembles the random baseline. However, we use the data likelihood to guide the random search for the generation of counterfactuals. We first generate a random seed of possible starting events ($\prob{e_0}$). Afterwards, we randomly sample subsequent events by iteratively sampling new activities according to the transition probabilities we gathered from the data ($\prod_1^T \cprob{e_t}{e_{t-1}}$). Given the sequence, we simply sample the features per event from $\cprob{f_t}{e_t}$. We call this baseline \textbf{Sample-Based} (SBGW). In contrast to both sampling-based baselines, the last baseline leverages actual examples of the data. We refer to this case-based approach as \textbf{Case-Based baseline} (CBGW). The idea is to randomly pick traces from the log and evaluate them using the viability measure.

\subsection{Experimental Setup}
\label{sec:experimental_setup}
% \subfile{content/sections/sec_experimental_setup}
% Counterfactual generation is notorious for lacking a standardised evaluation procedure. Nonetheless, we try to address our research questions with the following experiments. 
All the experiments were run on a Windows machine with 12 processor cores (Intel Core i7-9750H CPU 2.60GHz) and 32 GB Ram. The code is written in Python version 3.8. 
The models were developed with Tensorflow~\cite{abadi2016tensorflow} 
and NumPy~\cite{2020NumPy-Array}. 
We provide the full code and instructions on Github~\cite{hundogan_CREATEDGeneratingViable_2022}.

% \textbf{Experiment 1: Model Selection}
% The first set of simulations is dedicated to choosing among a subset of operator combinations and selecting appropriate hyperparameters. 
% First, we reduce the number of models that we compare against the baseline approaches in later experiments. In terms of operators, we introduced three initiators, three selectors, three crossers, two mutators and three recombiners. Hence, we compare all \NumEvoCombinations evolutionary operator combinations. We compute all possible configurations without changing any hyperparameter.

% After executing all preliminary simulations, we choose the best evolutionary generators and compare them with all baseline models in all subsequent experiments.

In terms of operators, we introduced three initiators, three selectors, five crossers, two mutators, and three recombiners. 
\textcolor{red}{
For the experiments, we exclude the random mutator as preliminary experiments showed that it often leads to results with a feasibility of 0.
} To reduce the number of model configurations, we initially compare all \NumEvoCombinations evolutionary operator combinations. We select the best three models and compare them to the three baseline models. 
Afterwards, we assess the viability of all the chosen evolutionary and baseline generators. \textcolor{red}{We sample 10 factuals from the BPIC-25 dataset and use our models as well as the baselines to generate 50 counterfactuals for each factual.} We determine the mean viability across the counterfactuals. We expect the evolutionary algorithms to outperform the baselines when it comes to viability.
In the end, we assess the quality of the generated counterfactuals. In line with \cite{hsieh_DiCE4ELInterpretingProcess_2021}, we aim to answer the question \emph{what would one have had to change in order to flip the outcome of a process}. The goal is to show that the counterfactuals our models generate are viable without having to rely on domain-specific knowledge. \textcolor{red}{In the current paper, we did not include any results of the individual viability components. Furthermore, we refer to \cite[p.64]{Hundogan2022} for more specific and extensive observations.}

\section{Results}
\label{ch:results}
% This section presents the results of each evaluation step. Furthermore, we analyse the results.

\subsection{Experiment 1: Comparing with Baseline Generators}
\label{sec:experiment2}
\begin{figure}[h]
    \centering
    \includegraphics[width=0.6\textwidth]{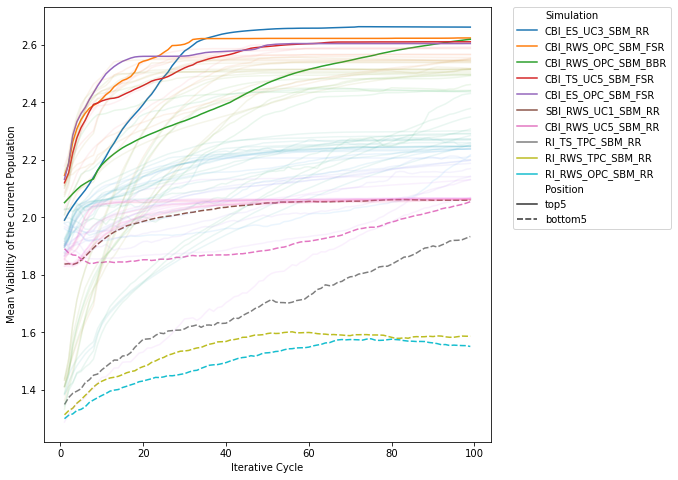}
    \caption{This figure shows the average viability of the five best and worst model configurations. The x-axis shows how the viability evolves for each evolutionary cycle. The semi-transparent lines are the model configurations that are neither in the best five nor worst five groups. They show the general trend of the viability improvement.}
    \label{fig:average-viability}
\end{figure}

We examined a set of model-configurations containing \NumEvoCombinations elements. We choose to run each model configuration for 100 evolution cycles. 
% For all model configurations, we use the same 4 factual \glspl{instance} randomly sampled from the test set. \autoref{fig:average-viability} shows the bottom and top-5 model configurations based on the viability after the final iterative cycle. We also show how the viability evolves for each iteration. 
\textcolor{red}{We randomly sample four factual \glspl{instance} from the test set. Afterwards, we use the average viability across the instances to evaluate all model configurations.} \autoref{fig:average-viability} shows the bottom and top-5 model configurations based on the viability after the final iterative cycle. The figure also shows how the viability evolves for each iteration. 

According to \autoref{fig:average-viability}, \emph{CBI-ES-UC3-SBM-RR}, \emph{CBI-RWS-OPC-SBM-BBR}, and \emph{CBI-RWS-OPC-SBM-FSR} are the best model configurations. As all best-performing model-configurations use the \emph{Case-Based Inititiation}-operator, we identify it as the most important configuration. The results suggest that the initiation operator governs the starting point of the optimisation.
For the following experiment, we ran each evolutionary algorithm for 200 iterative cycles and set the mutation rate to 0.01.  

\begin{figure}[b]
    \centering
    \includegraphics[width=0.8\textwidth]{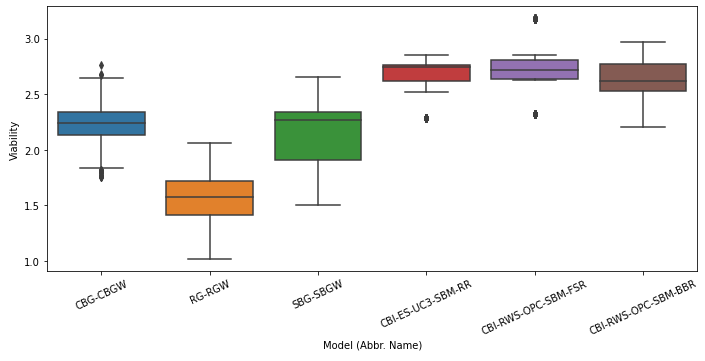}
    \caption{This figure shows boxplots of the viability of each model's generated counterfactuals.}
    \label{fig:exp4-winner}
\end{figure}

Next, we employed the baseline models mentioned in Section~\ref{sec:model_generation} and examined their results across all datasets. We randomly sampled 20 factuals from the test set and used the same factuals for every generator. We ensured that the outcomes are evenly divided. The remaining procedure followed the established practice of previous experiments. 
The results in \autoref{fig:exp4-winner} show that the evolutionary algorithm \optional{CBI-ES-UC3-SBM-RR} returns better results when it comes to the mean viability. The worst model is the randomly generated model. 
The Case-Based model appears to be evenly and normally distributed at a viability of \optional{2.25}. The \optional{CBI-RWS-OPC-SBM-FSR} has outliers that far exceed and underperform against other evolutionary algorithms on both ends.

\begin{figure}[t]
    \centering
    \includegraphics[width=0.8\linewidth]{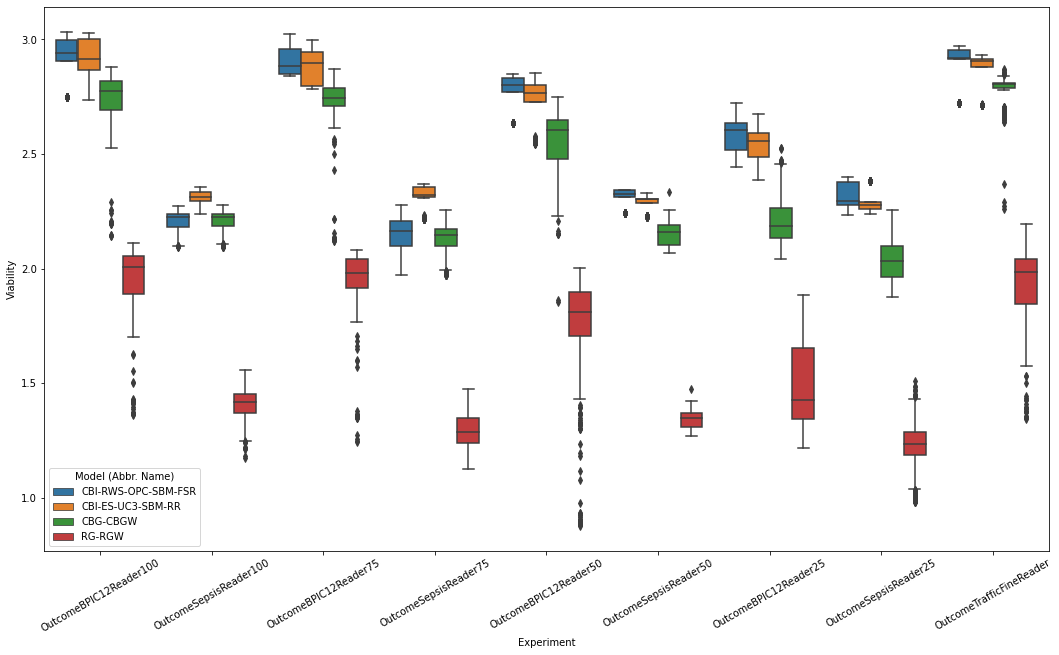}
    \caption{Boxplots of the viability of each model's generated counterfactuals across a heterogeneous collection of datasets.}
    \label{fig:exp5-winner}
\end{figure}

\autoref{fig:exp5-winner} displays the results of running each algorithm on a set of different datasets. The figure shows a clear dominance of the evolutionary models across all datasets. 
Here, \emph{CBI-ES-UC3-SBM-RR} and \emph{CBI-RWS-OPC-SBM-FSR} display a higher median of viability across all datasets. 
This is unsurprising as the evolutionary algorithm uses initiators based on the baselines. 
However, it is surprising that the evolutionary models consistently outperform the \ModelCBG (green) across all datasets. In six out of nine datasets, we see an improvement of at least 0.15. 
The highest median is reached for \emph{CBI-RWS-OPC-SBM-FSR} at 2.94. 
The \ModelRNG never manages to come even close to the case-based model. Except for the BPIC12-100 dataset, the \ModelRNG has a median below 2. 

The results for \autoref{fig:exp5-winner} show that both evolutionary algorithms outperform the competition across \emph{all} datasets and against \emph{all} baselines. 
This result shows that the algorithm can outperform baselines regardless of the process log and its length.
The baseline comparison also shows that we can optimise towards viability successfully. Recall that we defined four criteria for the viability of counterfactuals (similarity, sparsity, feasibility, and delta in likelihood); a model optimising towards those criteria can apparently return superior results.

\subsection{Experiment 2: Qualitative Assessment}
\label{sec:experiment4}
% \subsubsection{Results}
% \subfile{content/sections/sec_experiment_7}

\todo{XL: I would remove the resource column in table 4 and 5.}
\autoref{fig:exp7-FSR-1} shows the generation of the model-configuration \optional{CBI-RWS-OPC-SBM-FSR} and the model of \cite{hsieh_DiCE4ELInterpretingProcess_2021}. Both models also return reasonable counterfactuals. The counterfactual sequence of events of both approaches are almost identical. For instance, our counterfactual and the D4EL counterfactual recognize that after O-SENT, there appears at least one \emph{W-Completeren aanvraag} and one \emph{W-Nabellen offertes} that eventually leads to an acceptance of the counterfactual. We also see that both evolutionary algorithms start the process with the correct sequence of A-SUBMITTED, A-PARTLYSUBMITTED and A-PREACCEPTED. These are strictly the same across all cases. If our generative model had not recognised these, one could question its utility.

In \autoref{fig:exp7-FSR-2} we applied the same approach on a different dataset. The generator generates a counterfactual that is close to the original factual and only modifies the number of open cases. Here, we can conclude that a sudden increase in open cases during the \emph{Add penalty} step results in a change of outcome.

The examples show that our generative approach does not rely on domain knowledge, such as milestones. In contrast, the approach by \cite{hsieh_DiCE4ELInterpretingProcess_2021} only applies to datasets with clear milestones such as \emph{BPIC-12}.

\begin{table}[t]
    \centering    
    \caption{A comparison between the CBI-RWS-OPC-SBM-FSR and D4EL}
    \label{fig:exp7-FSR-1}
    \resizebox{\linewidth}{!}{
    \begin{tabular}{lllllllllll}
\toprule
\multicolumn{4}{l}{Factual Seq.} & \multicolumn{4}{l}{Our CF Seq.} & \multicolumn{3}{l}{DiCE4EL CF Seq.} \\
Amount & Activity & Outcome & Resource & Amount & Activity & Outcome & Resource & Activity & Resource & Amount \\
\midrule
5 000 & A-SUBMITTED & 0 & 112 & 7 000 & A-SUBMITTED & 1 & 112 &  &  &  \\
5 000 & A-PARTLYSUBMITTED & 0 & 112 & 7 000 & A-PARTLYSUBMITTED & 1 & 112 &  &  &  \\
5 000 & A-PREACCEPTED & 0 & 101 & 7 000 & A-PREACCEPTED & 1 & 112 &  &  &  \\
5 000 & W-Afhandelen leads & 0 & 101 &  &  &  &  & A-SUBMITTED & 112 & 5 000 \\
5 000 & A-ACCEPTED & 0 & 111 &  &  &  &  & A-PARTLYSUBMITTED & 112 & 5 000 \\
5 000 & O-SELECTED & 0 & 111 & 7 000 & A-ACCEPTED & 1 & 111 & A-PREACCEPTED & 112 & 5 000 \\
5 000 & A-FINALIZED & 0 & 111 & 7 000 & O-SELECTED & 1 & 111 & A-ACCEPTED & 1 & 5 000 \\
5 000 & O-CREATED & 0 & 111 & 7 000 & A-FINALIZED & 1 & 111 & O-SELECTED & 1 & 5 000 \\
5 000 & O-SENT & 0 & 111 & 7 000 & O-CREATED & 1 & 111 & A-FINALIZED & 1 & 5 000 \\
5 000 & W-Completeren aanvraag & 0 & 111 & 7 000 & O-SENT & 1 & 111 & O-CREATED & 1 & 5 000 \\
5 000 & W-Nabellen offertes & 0 & 111 & 7 000 & W-Completeren aanvraag & 1 & 111 & O-SENT & 1 & 5 000 \\
5 000 & O-CANCELLED & 0 & 111 &  &  &  &  & W-Completeren aanvraag & 1 & 5 000 \\
5 000 & A-CANCELLED & 0 & 111 & 7 000 & W-Nabellen offertes & 1 & 111 & O-SENT-BACK & 11259 & 5 000 \\
5 000 & W-Nabellen offertes & 0 & 111 & 7 000 & W-Nabellen offertes & 1 & 111 & W-Nabellen offertes & 11259 & 5 000 \\
 &  &  &  & 7 000 & O-ACCEPTED & 1 & 629 & O-ACCEPTED & 9 & 5 000 \\
\bottomrule
\end{tabular}
    }
\end{table}

\begin{table}[tb]
    \centering    
    \caption{A counterfactual for the Traffic-Fines dataset by the CBI-RWS-OPC-SBM-FSR model}
    \label{fig:exp7-FSR-2}
    \resizebox{\linewidth}{!}{
    \begin{tabular}{llllllll}
\toprule
\multicolumn{4}{l}{Factual Seq.} & \multicolumn{4}{l}{Our CF Seq.} \\
Open Cases & Activity & Outcome & Resource & Open Cases & Activity & Outcome & Resource \\
\midrule
16 318 & Create Fine & 1 & 537 & 15 742 & Create Fine & 0 & 537 \\
16 612 & Send Fine & 1 & 537 & 16 504 & Send Fine & 0 & 537 \\
16 693 & Insert Fine Notification & 1 & 537 & 16 693 & Insert Fine Notification & 0 & 537 \\
16 972 & Add penalty & 1 & 537 &  &  &  &  \\
\bottomrule
\end{tabular}
    }
\end{table}

% \begin{table}[h]
%     \centering    
%     \resizebox{\linewidth}{!}{
%     \input{./tables/counterfactuals/counterfactual_general-ES-EGW-CBI-ES-UC3-SBM-RR-IM-49-3.tex}
%     }
% \caption{A comparison between the CBI-ES-UC3-SBM-RR and D4EL}
% \label{fig:exp7-FSR-2}
% \end{table}

\subsection{Discussion and Limitations}

All models successfully flip the outcome of the prediction model and are close to the factual. In contrast, the model by \cite{hsieh_DiCE4ELInterpretingProcess_2021} proposes more changes to the sequence. It is important to recall that the generated counterfactuals focus on explaining the prediction model rather than the true process. More specifically, our generative model shows which events and attributes have to be present or omitted to flip the outcome of the prediction model. 
% If our framework attempts to explain how a prediction model behaves, then its applicability to real-world scenarios depends on the prediction model's performance. 

In contrast to \cite{hsieh_DiCE4ELInterpretingProcess_2021}, we show that we can create these counterfactuals without incorporating domain-specific knowledge, such as an understanding of milestone patterns. 
Domain knowledge can help to improve or evaluate our solutions. However, they are not strictly required.  
Furthermore, our models can generate sequences not present within the input event log. Case-based solutions often overlook this aspect, as they are heavily biased toward the input data. 

\textcolor{red}{It is worthwhile to discuss that \emph{counterfactual sequences} differ from \emph{counterfactual rules} or \emph{explanations}. To obtain explicit explanations or rules, the generated counterfactuals should be compared to the factual. Our framework enables some alignments between the generated counterfactuals with the factual sequence (see~\autoref{fig:exp7-FSR-1}), which may act as an explanation. We consider deriving rules as a post-prior analysis, which is interesting for future work.}

Our viability components showed that they can lead to an optimised solution. However, there are most likely other ways to operationalise viability criteria. In addition, what makes an excellent counterfactual and how we can quantify that is still a subject of debate. Currently, there is a lack of standardized evaluation protocols, benchmark techniques, and datasets. As a result, many researchers fall back on defining their custom evaluation methods.
% as there is no standardized way to evaluate the viability of a counterfactual. 
In fact, this is still an open research question~\cite{hsieh_DiCE4ELInterpretingProcess_2021,mothilal_ExplainingMachineLearning_2020}. Therefore, we often have to evaluate the counterfactuals in some subjective and qualitative way. In this paper, we decided to compare the counterfactuals with another approach in the literature and the factual themselves. Because our counterfactuals produced reasonable results, we deemed them viable. As future work, we also see value in incorporating experts to evaluate such an approach.

\section{Conclusion}
\label{ch:conclusion}
% \subfile{content/sections/sec_conclusion}
In this paper, we proposed CREATED, a modular framework to generate viable counterfactuals.
The framework incorporates an evolutionary algorithm to generate counterfactual sequences while not requiring any domain knowledge other than the log itself. In addition, we proposed a viability measure to quantify and assess the quality of counterfactual sequences when compared to a factual sequence. The viability measure takes four aspects into account: feasibility, the delta in flipping the outcome prediction, similarity, and sparsity. The approach is capable of generating counterfactuals without explicit knowledge about the domain, as we only require the log. We achieve this by incorporating a Markov model trained on the event log. Our evaluation shows that our framework can generate counterfactual sequences which are higher than our naive baselines (i.e., case-based, sample-based, and random baselines). With these results, we demonstrate that optimizing a viability measure does generate higher-quality counterfactuals. We also compared the generated counterfactuals to the state-of-the-art method in the literature and show that our framework can generate similar counterfactuals, without using domain knowledge. 
The current feasibility measure tends to return lower values than other viability components as it is very sensitive to trace length. In the future, we aim to investigate better feasibility measures. 
\bibliographystyle{splncs04}
% \bibliographystyle{splncs04nat}
% \bibliography{./references/bibliography.bib}
\bibliography{./ref.bib}




% LNCS Homepage, \url{http://www.springer.com/lncs}. Last accessed 4
% \end{thebibliography}

% XIXI Comments on Structure
% XIXI: After the architecture follow with the evolutionary framework and then viability function.
% XIXI: LSTM-Prediction model into the architecture section.

% XIXI
% Intro 3-4 pages
% Background 0.5 pages
% Approach 5 pages
% - Architecture 1 pages
% - Evolutionary 2 pages
% - Viability 2 page
% Evaluation 5 pages

% DISCUSSION
% First repeat parts of 6.1 and 6.2 again.
% Emphasize again that we do not need domain
% Link to the feasibility 
% Also talk about the casebased inititator - DONE
% 
% FINAL TODOS
% Add oxford commas
% Add the github link

\end{document}